# A Web-based Mpox Skin Lesion Detection System Using State-of-the-art Deep Learning Models Considering Racial Diversity

Shams Nafisa Ali[1, ‡], Md. Tazuddin Ahmed[1, ‡], Tasnim Jahan[1, §], Joydip Paul[1, §], S. M. Sakeef Sani[1], Nawsabah Noor[2], Anzirun Nahar Asma[2], Taufiq Hasan[1, 3], *Senior Member, IEEE*

*Abstract* — The recent 'Mpox' outbreak, formerly known as 'Monkeypox', has become a significant public health concern and has spread to over 110 countries globally. The challenge of clinically diagnosing mpox early on is due, in part, to its similarity to other types of rashes. Computer-aided screening tools have been proven valuable in cases where Polymerase Chain Reaction (PCR) based diagnosis is not immediately available. Deep learning methods are powerful in learning complex data representations, but their efficacy largely depends on adequate training data. To address this challenge, we present the "Mpox Skin Lesion Dataset Version 2.0 (MSLD v2.0)" as a follow-up to the previously released openly accessible dataset, one of the first datasets containing mpox lesion images. This dataset contains images of patients with mpox and five other non-mpox classes (chickenpox, measles, hand-foot-mouth disease, cowpox, and healthy). We benchmark the performance of several state-of-the-art deep learning models, including VGG16, ResNet50, DenseNet121, MobileNetV2, EfficientNetB3, InceptionV3, and Xception, to classify mpox and other infectious skin diseases. In order to reduce the impact of racial bias, we utilize a color space data augmentation method to increase skin color variability during training. Additionally, by leveraging transfer learning implemented with pre-trained weights generated from the HAM10000 dataset, an extensive collection of pigmented skin lesion images, we achieved the best overall accuracy of $83.59 \pm 2.11\%$. Finally, the developed models are incorporated within a prototype web application to analyze uploaded skin images by a user and determine whether a subject is a suspected mpox patient.

## I. INTRODUCTION

The global outbreak of the virus previously called 'Monkeypox', now referred to as mpox, has caused widespread concern over the past year and continued to be a major topic in public health news headlines. In July 2022, the World Health Organization (WHO) declared it a Public Health Emergency of International Concern (PHEIC) due to the significant risk associated with the virus [1]. The World Health Network (WHN) has also emphasized the need for coordinated global action to combat the spread of the disease, given its potential for deadly outcomes [2]. Recent epidemiological data indicate that the mpox outbreak is slowing down in the American and European regions, while the transmission is still ongoing in African regions [3]. In May 2023, WHO announced that mpox is no longer classified as a PHEIC [4]. Despite this, the study of mpox remains relevant, as there is a looming threat of another possible multi-country outbreak.

‡ These authors share first authorship on.

§ These authors share second authorship on and contributed equally.

[1] Shams Nafisa Ali, Md. Tazuddin Ahmed, Tasnim Jahan, Joydip Paul, S. M. Sakeef Sani and Taufiq Hasan are with mHealth Lab, Department of Biomedical Engineering, BUET, Bangladesh. Email: taufiq@bme.buet.ac.bd.

[2] Nawsabah Noor and Anzirun Nahar Asma are with Popular Medical College, Dhaka, Bangladesh.

[3] Taufiq Hasan has a secondary affiliation with Center for Bioengineering Innovation and Design, Department of Biomedical Engineering, Johns Hopkins University, Baltimore, MD.

Mpox is a dsDNA virus that originates from the *Poxviridae* family and *Orthopoxvirus* genus [5]. Monkeys, rodents, squirrels, non-human primates, and several other animal species have been identified as primary vectors for transmission [5]. Since its first confirmed human case, in 1970, in the Democratic Republic of Congo (DRC), the human-to-human transmission of mpox has come to notice and is marked as endemic in the tropical rainforest region of Africa [5]. Since January 2022, mpox has been reported in 110 member states from six WHO regions, with 86,496 laboratory-confirmed cases and 111 deaths reported as of March 13, 2023 [3]. Most of these cases have been reported in nations with no history of mpox transmission [6]. While mpox cases peaked during July-August 2022 and have since declined, it is crucial to establish frameworks that can be readily applied to diagnosis and screening of mpox in case of its reappearance, given the recurrence of health threats by viral species responsible for communicable diseases, such as SARS-CoV, MERS-CoV, and SARS-CoV2 [7], [8].

Mpox disease symptoms closely resemble the rashes of other diseases such as chickenpox, measles, rickettsial pox, smallpox, and hand-foot-mouth disease [5]. From case reports and demographics, it also appears to be a relatively rare disease among Asians, native Hawaiians, and other Pacific Islanders [9]. These factors, along with the inadequate Polymerase Chain Reaction (PCR) testing facilities in many countries, pose significant challenges for healthcare professionals. In this scenario, AI-based automated computer-aided systems may substantially contribute to resolving the core impediments in the way of rapid and accurate initial screening of mpox.

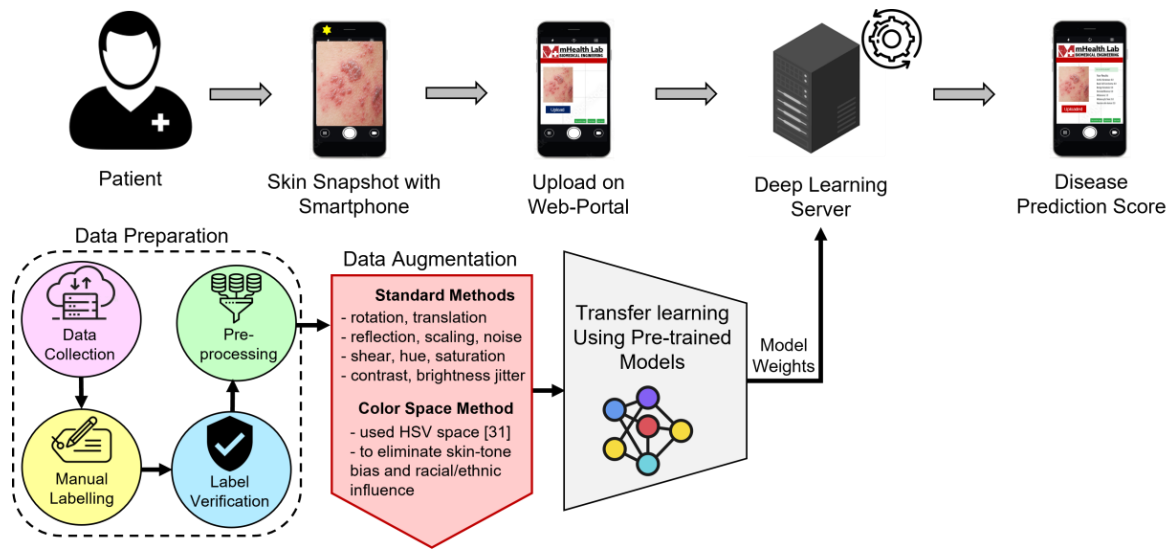

Fig. 1: A flow-diagram of the proposed mpox detection system. A prototype web-app is developed that incorporates the best-performing deep learning model to detect mpox from skin lesions uploaded by users.

In recent years, the multi-faceted applications of deep learning (DL), particularly the variations of Convolutional Neural Networks (CNNs), have revolutionized different fields of medical science due to their superior learning capability compared to conventional machine learning techniques [10], [11]. When trained with ample data, these networks can automatically extract salient features from images, creating optimal representations for specified tasks [12], [13]. However, the necessity for large amounts of data and time-consuming training with specialized computational resources hinders the applicability of DL-based frameworks [14]. While accelerators (e.g., GPU, TPU) can resolve time and resource-related issues, obtaining unbiased and homogeneous clinical data remains challenging. One well-known method of increasing dataset size is data augmentation [15], which generates additional samples through slight modifications of existing data. In cases of data scarcity, transfer learning [14] is commonly used, where a CNN model pre-trained on a large dataset (e.g., ImageNet [16]) transfers knowledge for context-specific learning on a smaller dataset.

Inspired by the superior performance of the DL algorithms across different domains, research groups worldwide have attempted to create datasets containing mpox skin lesion images to train effective learning algorithms [17]–[20]. Our research group was one of the first to release a dataset with mpox lesion images, "Monkeypox Skin Lesion Dataset (MSLD)"[1], containing web-scrapped images of patients (validated through Google's Reverse Image Search) categorized into two classes: 'monkeypox' and 'non-monkeypox' (measles, chickenpox). However, several issues plague most of these datasets, including ours: a few images had mislabels due to the lack of professional scrutiny by dermatologists, and some images were of poor quality with watermarks and distortions. Although the instant release of such unscrutinized datasets was essential during the initial surge of mpox cases, now that the cases have come under control, ensuring the clinical soundness of the data through expert verification and feedback incorporation is crucial for developing effective algorithms to tackle any future cases of monkeypox or diseases with similar skin lesions.

In this paper, we introduce an updated version of our previously released dataset, "Mpox Skin Lesion Dataset Version 2.0 (MSLD v2.0),"[2] a publicly available dataset consisting of web-scraped images of patients with mpox and non-mpox cases, including chickenpox, measles, cowpox, hand-foot-mouth disease, and healthy individuals. The images are taken from different body parts, such as the face, neck, hand, arm, and leg. Our preliminary study explores the potential of deep learning models for the early detection of mpox disease, leveraging transfer learning on various architectures, including VGG16 [21], ResNet50 [22], DenseNet121 [23], MobileNetV2 [24], EfficientNetB3 [25], InceptionV3 [26] and Xception [27]. Furthermore, we have developed a web application[3] using the open-source Streamlit framework that analyzes uploaded images and predicts whether the subject is a suspected mpox patient requiring urgent consultation with a physician. Our current working pipeline is illustrated in Fig. 1.

"Skin color bias," also known as the "white lens problem", has long been an issue in AI-based diagnosis for skin lesion images [13]. Most benchmark datasets, such as HAM10000 [28], ISIC challenge dataset 2018 [29], and PH2 (Prado Hospital 2) dataset [30], consist primarily of images of individuals with lighter skin tones, leading to inaccuracies in diagnoses for individuals with darker skin tones. However, mpox is prevalent in African regions; consequently, most of the mpox data come from dark-skinned individuals, potentially introducing a bias opposite to the "white lens problem." Additionally, the non-mpox classes had a comparatively higher number of samples with lighter skin tones. To reduce the impact of skin tone bias in our dataset, we adopted a recently proposed skin-color agnostic color-space augmentation before classification [31]. This technique simulates the effect of a diverse range of skin tones and ethnicities, helping to mitigate any bias in our dataset.

The main contributions of this paper are summarized below:

- We introduce the Mpox Skin Lesion Dataset Version 2.0 (MSLD v2.0) containing web-scrapped skin lesion images of mpox and non-mpox patients.
- We explore the potential of DL-based models, including VGG16, ResNet50, DenseNet121, MobileNetV2, EfficientNetB3, InceptionV3, and Xception architectures for early detection and screening of mpox from skin lesion images.
- We adopt a skin-color agnostic color-space augmentation method to improve the skin tone variability in the training dataset and thus increase the generalizability of the model to various patient ethnicities.
- We developed a web-app capable of predicting whether a subject is a potential mpox suspect and should consult a physician or not.

The remainder of the paper is organized as follows. In Sec. II, we present a brief background on different stages of mpox lesions. Sec. III contains a brief review of the relevant literature. Sec. IV provides a detailed description of dataset development and its acquisition procedure. Sec. V outlines the experiments performed on the dataset and the results associated with this. The web-app description is presented in Sec. VI. Finally, Sec. VII and VIII summarize this work's contributions and discuss future directions.

## II. BACKGROUND

Mpox is a typically self-limited disease, with symptoms lasting between 2 to 4 weeks. The severity of the disease generally depends on the extent of virus exposure, the health status of the patient, and the complications. The disease tends to affect children more severely. The virus has an incubation period that ranges from 5 to 21 days [5]. During the invasion period (which lasts from 0 to 5 days), patients commonly experience symptoms such as fever, lymphadenopathy (swollen lymph nodes), myalgia (muscle ache), asthenia (physical weakness), and severe headache. The rash begins within 1-3 days of fever onset and is usually noticed on the face, palms of the hands, and soles of the feet [5]. In the skin eruption phase (2-4 weeks), the lesions follow a four-stage progression: macules (lesions with a flat base) develop into papules (raised, firm, and painful lesions), which then become vesicles (filled with clear fluid) and finally pustules (filled with pus) before encrustation. Consequently, the lesions may appear slightly different as they progress through these stages (see Fig. 2).

## III. RELATED WORK

Classifying different types of skin lesions is a challenging problem due to high inter-class similarity and intra-class variability [32]. In addition, the lack of available skin lesion data has been the primary challenge in developing deep-learning models for mpox detection. Thus, when the mpox cases surged in 2022, researchers mainly focused on developing a reliable mpox skin image dataset with several non-mpox classes that have high similarity with mpox in terms of the appearance and characteristics of the rashes. After several such web-scrapped datasets were released, studies on bench-marking and developing deep learning architectures for mpox detection emerged. Thus, our literature survey concentrates on two primary areas: (i) the development of mpox skin lesion datasets and (ii) the evolution of classification techniques.

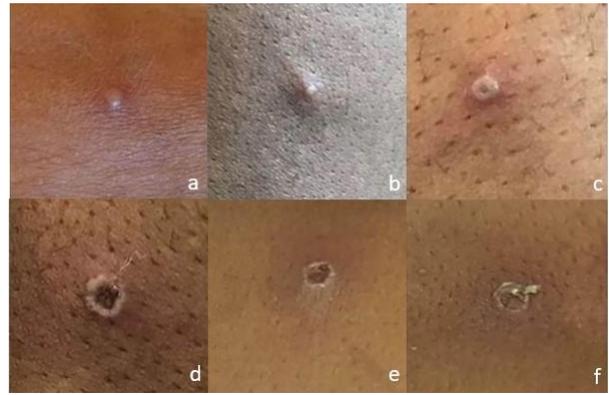

Fig. 2: The mpox lesion through its various stages: (a) early vesicle, (b) small pustule (⌀ 2mm), (c) umbilicated pustule (⌀ 3-4mm), (d) ulcerated lesion (⌀ 5mm), (e) crusting of a mature lesion, (f) partially removed scab.

In terms of dataset development, as mentioned in Sec. I, we have previously introduced the first version of our mpox dataset as MSLD v1.0 [33], which contains images belonging to two classes: 102 'mpox' images and 126 'non-mpox' (chickenpox and measles) images. During the same period, Ahsan *et al.* published one of the first few datasets on mpox skin lesion images [17]. The dataset contains skin images of 4 classes: 43 mpox, 47 chickenpox, 17 measles, and 54 healthy. Subsequently, the "Monkeypox Skin Images Dataset (MSID)" was released on Kaggle consisting of images from the same four classes, i.e., mpox (279 images), chickenpox (107 images), measles (91 images), and healthy (293 images) [19]. Immediately after the release of MSID, Islam *et al.* published another dataset, which was previously available on Kaggle, containing images labeled for six classes: mpox (117 images), chickenpox (178 images), smallpox (358 images), cowpox (54 images), measles (47 images), and healthy (50 images) [18]. However, none of these datasets were verified by a dermatologist, which increases the risk of mislabeling images, leading to erroneous disease identification. Several other limitations include the presence of watermarks, poor aspect ratio, and irrelevant images due to improper filtering in the web-scraping process. It is also important to note that the 'smallpox' class in the dataset released in [18] is error-prone since smallpox is now extinct. A few other datasets are also created using our dataset MSLD v1.0 as the foundation and then adding to it [20].

Various classification studies were conducted using the pre-liminary version of MSLD and other available datasets. Situala *et al.* performed classification using 13 DL models pretrained on ImageNet weights [34]. They proposed an ensemble on Xception and DenseNet-169 based on performance. The authors also explained the performance of their best-performing model, Xception, using Gradient-weighted Class Activation Mapping (Grad-CAM) and Local Interpretable Model-Agnostic Explanations (LIME). Abdelhamid *et al.* proposed two algorithms for improved classification results [35]. The first uses GoogleNet architecture based on transfer learning for feature extraction, and the second approach consists of a binary hybrid algorithm for feature selection and a hybrid algorithm for optimizing the neural network's parameters. The Al-Biruni Earth radius algorithm, the sine-cosine algorithm, and the particle swarm optimization algorithm are examples of meta-heuristic optimization algorithms that are used for feature selection and parameter optimization. Ahsan *et al.* proposed a modified VGG16 model for classification and explained the feature extraction of the model using LIME [36]. Islam *et al.* used 5-fold cross-validation to fine-tune seven DL models with pre-trained weights from ImageNet on their dataset [37]. Alakus *et al.* constructed a deep learning algorithm to categorize the DNA sequences of the MPV and HPV viruses that cause mpox and warts, respectively [38]. The findings revealed an average accuracy of 96.08% and an F1-score of 99.83%, demonstrating that the two diseases can be correctly identified based on their DNA sequences. Sahin *et al.* created an Android mobile app that uses deep learning to aid in detecting mpox using the DL architectures EffcientNetb0 and MobileNetv2 [39]. Haque *et al.* attempted to integrate deep transfer learning-based methods and a convolutional block attention module (CBAM) to focus on the relevant portion of feature maps to conduct an image-based classification of mpox [40]. They used the DL architectures VGG19, Xception, DenseNet121, EfficientNetB3, and MobileNetV2. Their proposed model, XceptionCBAM-Dense, was reported to achieve 83.89% accuracy on our dataset. Kumar investigated various deep CNN models with multiple machine learning classifiers for mpox disease diagnosis [41]. For this, bottleneck features of three CNN models, i.e., AlexNet, GoogleNet, and VGG16, are explored with multiple machine learning classifiers such as SVM, KNN, Naïve Bayes, Decision Tree, and Random Forest. Yang *et al.* introduced an AI-based mpox detector primarily aimed at handling images taken from resource-constrained devices [20].

## IV. Dataset Preparation

During the initial peak outbreak phase of mpox, there was no publicly available dataset for the detection of mpox. Therefore, for the initial feasibility analysis of the AI-based mpox screening system, images of different body parts (face, neck, hand, arm, leg) of patients with mpox and non-mpox (measles, chickenpox) cases were collected from publicly available case reports, news portals, and reliable websites via web-scrapping [42] by our research group and the first version of "Mpox Skin Lesion Dataset (MSLD)" was released.

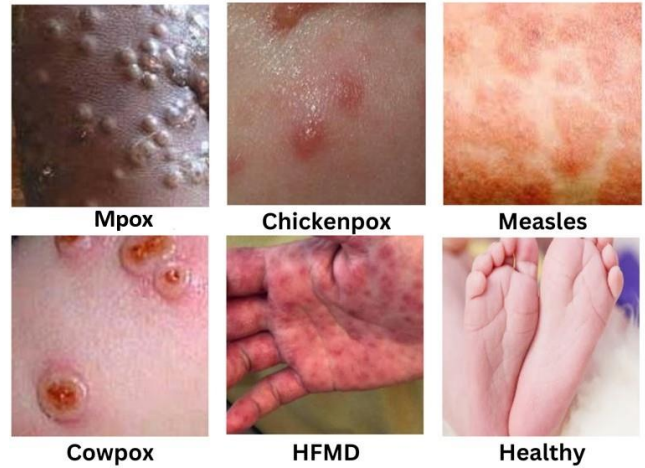

**Fig. 3:** Sample images of each class collected from the dataset.

### A. Data Collection Procedure

This current study presents an improved and expanded version of our dataset "Mpox Skin Lesion Dataset Version 2.0 (MSLD v2.0)". The MSLD v2.0 contains web-scraped images belonging to 6 classes: mpox (284 images), chickenpox (75 images), measles (55 images), cowpox (66 images), hand-foot-mouth disease or HFMD (161 images), and healthy (114 images) for multi-class classification. This retrospective data collection study was approved by our Institutional Review Board (IRB)[4] where the informed consent requirement was waived. In most cases, the patient's personal identifying information (including patient history, health, epidemiological information, and co-morbidity) was not disclosed in the original source of the image and thus was confidential. We acknowledge that web-scraped images could be from various sources, including copyrighted data as well as non-copyrighted, freely usable, reusable, and redistributed public-domain data [43]. Since MSLD v2.0 images are intended to be used only for research purposes, our IRB approved the data collection study under the "fair use" principle provided that the sources are appropriately listed and credited [44]. Our data collection protocol also included a prospective study component for collecting images from mpox patients in our IRB-approved clinical study sites. However, since there were no known cases of mpox in Bangladesh, the prospective study component could not be conducted.

### B. Dataset Screening

The collected skin images were processed through a 2-stage screening process. First, the out-of-focus, low-resolution, and low-quality images were discarded, and only the unique images that satisfy the quality requirements were selected. Next, the images were cropped to their region of interest and resized to 224 × 224 pixels while maintaining the aspect ratio. Finally, an expert dermatologist verified the disease label of each skin image. Fig 3 shows a few image samples from the dataset. A detailed distribution of the dataset is provided in Table I.

---

[4]This study was approved by Ethical Review Committee of Popular Medical College (Ref: PMC/Ethicalrc/2023/02).

TABLE I: Class distribution of the presented Mpox Skin Lesion Dataset (MSLD) v2.0

| Class label | No. of Original Images | No. of Unique Patients |
|---|---|---|
| Mpox | 284 | 143 |
| Chickenpox | 75 | 62 |
| Measles | 55 | 46 |
| Cowpox | 66 | 41 |
| Hand, foot and mouth disease | 161 | 144 |
| Healthy | 114 | 105 |
| Total | 755 | 541 |

## C. Standard Data Augmentation

In the next stage, to assist in the classification task and improve the generalizability of the learning algorithms, several data augmentation methods, including rotation, translation, reflection, shear, hue, saturation, contrast and brightness jitter, noise, and scaling, were applied to the dataset. Post-augmentation, the number of images increased by approximately 14-fold. These augmented images are also provided in a separate folder in the dataset to ensure the reproducibility of results.

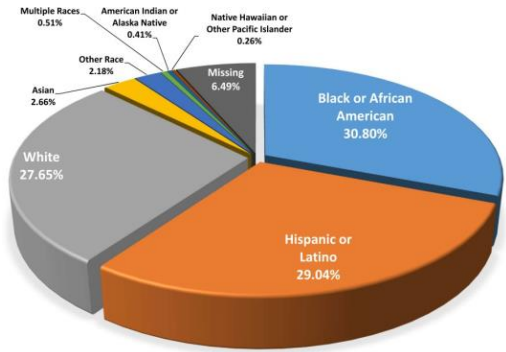

Fig. 4: Distribution of reported mpox cases by race and ethnicity.

## D. Color Space Augmentation

The prevalence of skin-tone bias in skin image datasets for automated skin detection is a well-known phenomenon, largely due to the imbalance in the distribution of training samples that tend to represent lighter skin tones [13]. This bias is also This bias is also evident in our dataset, as most of the images within classes such as Chickenpox, Cowpox, Measles, and HFMD primarily depict individuals with lighter skin tones. In contrast, the mpox class contains a mixture of individuals with both dark and light skin tones. A recent report from the Centers for Disease Control and Prevention (CDC) highlights the distribution of mpox cases among different racial and ethnic groups, thereby providing further evidence of this imbalance [45]. According to the report, out of all the reported mpox cases, 30.08% are identified as Black or African American, 29.04% as Hispanic or Latino, and 27.65% as White. The distribution of reported mpox cases by race and ethnicity is visually represented in Fig 4.

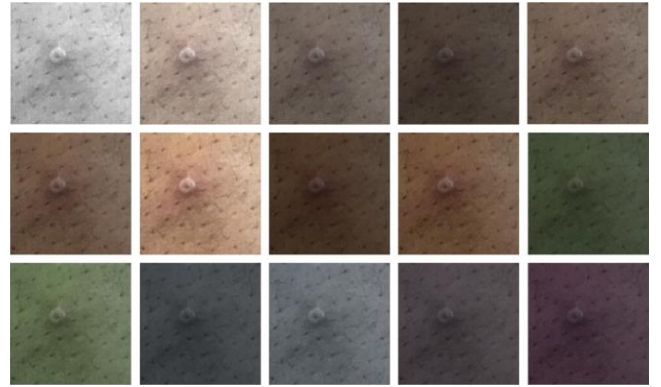

Fig. 5: Example of color space augmentation with varying HSV parameters on an mpox skin lesion image.

Studies have shown that the over-reliance on color information in skin disease detection techniques has imposed limitations due to skin-tone bias in the dataset. To address this issue, we also utilize color space augmentation in the HSV (hue, saturation, and value) space [31]. This augmentation aims to increase the universality of the dataset and reduce racial bias by transforming each image into 180 different versions with varying values in the HSV space. Doing so suppresses the reliance on color cues, and the training procedure is guided toward visual texture and context features. Alg. 1 describes our color space augmentation method.

---

**Algorithm 1:** Color Space Augmentation

**Input** : $X_t = \{x^i\}_{i=1,...,n}$ - Images from the dataset

**Output**: $X_{aug}$ - Augmented image dataset

1 Initialize $X_{aug}$=[]
2 Initialize $X_{hue}$=[]
3 Initialize $X_{sat}$=[]
4 Initialize $X_{val}$=[]
5 **for** $i \in \{1, 2, ..., n\}$ **do**
6   **for** $j \in \{H_1, H_2, ..., H_{final}\}$ **do**
7     $X_{hue} \Leftarrow$ hue_shift$(x^i, j)$
8     **for** $k \in \{S_1, S_2, ..., S_{final}\}$ **do**
9       $X_{sat} \Leftarrow$ saturation_scaling$(X_{hue}, k)$
10       **for** $l \in \{V_1, V_2, ..., V_{final}\}$ **do**
11         $X_{val} \Leftarrow$ value_scaling$(X_{sat}, l)$
12         $X_{aug} \Leftarrow (X_{aug}) \bigcup (X_{val})$
13         Reset $X_{val}$=[]
14       **end**
15       Reset $X_{sat}$=[]
16     **end**
17     Reset $X_{hue}$=[]
18   **end**
19 **end**
20 **return** $X_{aug}$

---

TABLE II: Performance comparison of different DL models using ImageNet Pre-trained weights for transfer learning. The best two results are shown in red and blue colors.

| Network | Accuracy (%) | Precision | Recall | F1 score |
|---------|-------------|-----------|--------|----------|
| VGG16 | 75.22± 3.16 | 0.79 ± 0.02 | 0.71 ± 0.06 | 0.72 ± 0.05 |
| ResNet50 | 77.94± 3.87 | 0.79 ± 0.05 | 0.76 ± 0.07 | 0.76 ± 0.06 |
| DenseNet121 | **81.70± 5.39** | 0.83 ± 0.04 | 0.79 ± 0.06 | 0.80 ± 0.06 |
| MobileNetV2 | 76.98± 4.65 | 0.81 ± 0.06 | 0.74 ± 0.05 | 0.75 ± 0.05 |
| EfficientNetB3 | 74.61± 3.94 | 0.75 ± 0.06 | 0.71 ± 0.06 | 0.72 ± 0.06 |
| InceptionV3 | 76.31± 2.66 | 0.79 ± 0.04 | 0.72 ± 0.03 | 0.74 ± 0.03 |
| Xception | 75.74± 6.20 | 0.75 ± 0.07 | 0.74 ± 0.08 | 0.73 ± 0.07 |

## V. EXPERIMENTS AND RESULTS

### A. Experimental Design

Our experimental evaluations are conducted using a five-fold cross-validation framework. The original images are divided into train, validation, and test sets, keeping an approximate distribution of 70:20:10 while preserving patient independence. Only images from the train and valid set were augmented. The experiments were subdivided into two separate studies. Only the standard augmentation techniques were applied for the first study, as stated in the previous section. For the second study, color space augmentation was performed as well in combination with standard augmentation methods, and changes in the results were investigated. We employed accuracy, precision, recall/sensitivity, and F1-score as our performance metrics.

### B. Pre-trained Networks and Transfer Learning

To evaluate the performance of DL-based classification algorithms on our MSLD v2.0 data, we have selected seven well-known CNN architectures- VGG16 [21], ResNet50 [22], DenseNet121 [23], MobileNetV2 [24], EfficientNetB3 [25], InceptionV3 [26] and Xception [27] pre-trained on the ImageNet dataset. We select these models as they have demonstrated excellent classifications through transfer performance in various computer vision and medical image analysis tasks. Transfer learning involves training a DL model on a large dataset, known as the 'source dataset', and then utilizing the learned model parameters to initialize training on a relatively smaller 'target dataset'.

In many cases, ImageNet pre-trained models perform satisfactorily well while using transfer learning for image-based classification tasks. However, we hypothesize that model performance may further improve for mpox detection if, instead of ImageNet data, the network is pre-trained using a large skin lesion image dataset. Therefore, to test this hypothesis, we also pre-trained our model using HAM10000 [28], a large open-access skin-lesion dataset. This dataset contains 10,015 images from seven categories of skin lesions, including Melanoma, Melanocytic Nevi, Basal Cell Carcinoma, Actinic Keratosis and Inta-Epithelial Carcinoma, Benign Keratosis, Dermatofibroma, and Vascular Lesions. Experimental results are discussed in the following sections.

### C. Implementation Details

Input images with dimensions (224, 224, 3) were fed into the selected pre-trained models. The fully connected layers were removed. We kept all the layers trainable. Next, we flatten the backbone model's output, followed by three blocks of fully connected (FC) layers, and dropout to the network. The FC layers had successively 4096, 1072, and 256 nodes while the dropout factors were respectively 0.3, 0.2 and 0.15. Finally, an FC layer with six nodes was employed with a softmax activation function for this multi-class classification task.

The network architectures were implemented in Keras and were accelerated using Nvidia K80 GPUs provided by Kaggle notebooks. The batch size was set to 16. The adaptive learning rate optimizer (Adam) with an initial learning rate of $10^{-5}$ and categorical cross-entropy loss function was employed for training

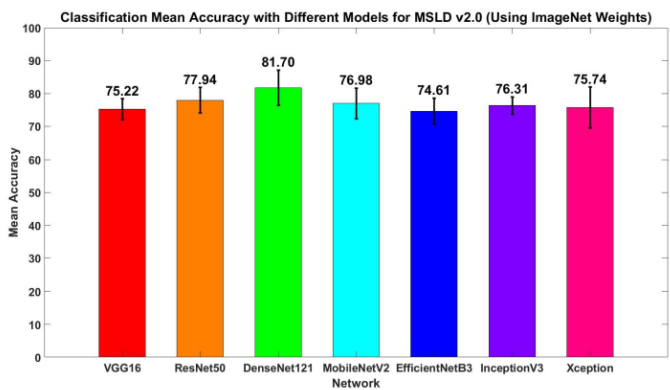

Fig. 6: Classification results with different DL models for transfer learning using ImageNet pre-trained weights.

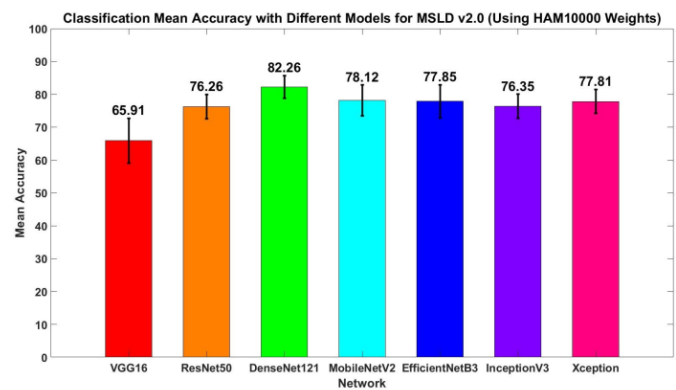

Fig. 7: Classification results with different DL models for transfer learning using HAM10000 pre-trained weights.

TABLE III: Performance comparison of different deep learning models using HAM10000 pre-trained weights for transfer learning. The best two results are shown in <span style="color:red">red</span> and <span style="color:blue">blue</span> colors.

| Network | Accuracy (%) | Precision | Recall | F1 score |
|---|---|---|---|---|
| VGG16 | 65.91 ± 6.82 | 0.71 ± 0.14 | 0.6 ± 0.06 | 0.62 ± 0.08 |
| ResNet50 | 76.26 ± 3.71 | 0.80 ± 0.04 | 0.74 ± 0.05 | 0.75 ± 0.04 |
| <span style="color:red">DenseNet121</span> | <span style="color:red">**82.26 ± 3.46**</span> | <span style="color:red">0.85 ± 0.05</span> | <span style="color:red">0.80 ± 0.07</span> | <span style="color:red">0.83 ± 0.05</span> |
| <span style="color:blue">MobileNetV2</span> | <span style="color:blue">78.12 ± 4.69</span> | <span style="color:blue">0.81 ± 0.04</span> | <span style="color:blue">0.75 ± 0.08</span> | <span style="color:blue">0.76 ± 0.06</span> |
| EfficientNetB3 | 77.85 ± 5.06 | 0.78 ± 0.09 | 0.79 ± 0.11 | 0.77 ± 0.10 |
| InceptionV3 | 76.35 ± 3.70 | 0.78 ± 0.06 | 0.73 ± 0.07 | 0.74 ± 0.06 |
| Xception | 77.81 ± 3.66 | 0.77 ± 0.04 | 0.75 ± 0.06 | 0.75 ± 0.06 |

TABLE IV: Comparison of the best-performing model using color space data augmentation on mpox skin lesion detection

| Model | Data augmentation | Accuracy (%) | Precision | Recall | F1 score |
|---|---|---|---|---|---|
| DenseNet121 | Standard | 82.26 ± 3.46 | 0.85 ± 0.05 | 0.80 ± 0.07 | 0.83 ± 0.05 |
| DenseNet121 | Standard + color-space | **83.59 ± 2.11** | 0.84 ± 0.02 | 0.85 ± 0.02 | 0.82 ± 0.03 |

### D. Comparing of Model Initialization Methods

In these experiments, the transfer learning models were initialized using two different approaches. The first employed pre-trained weights from ImageNet data, while the second utilized the HAM10000 skin lesion dataset. The results for the ImageNet-pre-trained models are summarized in Table II and Fig 6. These results show that DenseNet121 yields the best accuracy (81.70 ± 5.39%) while ResNet50 also shows competitive performance (77.94 ± 3.87%). In the second approach, we pre-trained our models using the HAM10000 skin image dataset and used these model parameters to initialize the transfer learning. The results are summarized in Table III and Fig 7. As anticipated, the performance metrics of most of the architectures have improved. The best-performing architecture in the first approach, DenseNet121, yielded 82.26% accuracy while transfer learning was performed with HAM10000 weights. Moreover, the revised initialization technique reduced the standard deviation in terms of the accuracy metric to 3.46%, indicating its consistency in performance across the five folds.

### E. Evaluation of Color Space Augmentation

In this second study, we aim to compare the performance between training our models using the data augmentation using standard and color space methods as discussed in sections IV-C and IV-D. Only the best-performing architecture from the previous study, DenseNet121, was used in this experiment. In the newly synthesized data using color space augmentation, each class had images from varying skin tones which could contribute to the learning procedure of the architecture. In Table IV, the performances of the best-performing models are compared to evaluate the impact of the color space augmentation in this mpox skin disease detection task. The results show that averaged accuracy across the five folds increased to 83.59 ± 2.11% from 82.26 ±3.46%. Concurrently, the standard deviation of the accuracy decreased, indicating the improved consistency of the models trained using color-space- augmented data. Color space augmentation also improved the recall/sensitivity from 0.80±0.07 to 0.85±0.02, whereas precision and F-1 score were relatively unchanged.

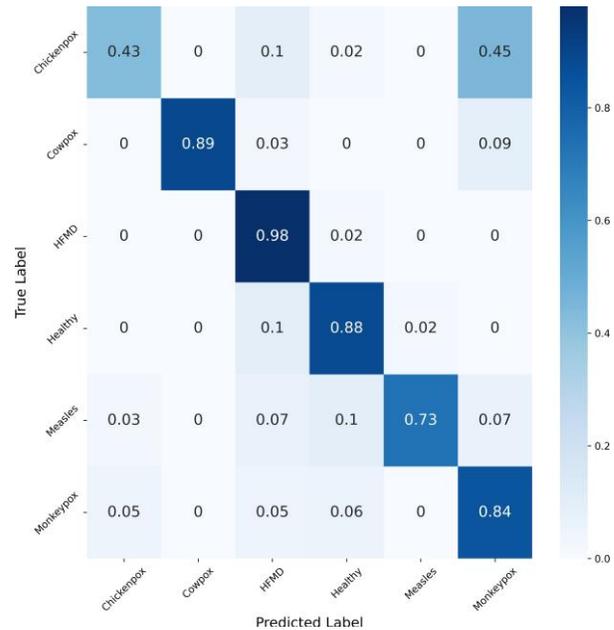

Fig. 8: Normalized Confusion Matrix depicting the classification performance across all folds.

To gain a comprehensive understanding of the overall performance across the dataset, we aggregated the individual confusion matrices for each fold by summing them together. This enabled us to consider and combine the results from all five folds. Then the summed confusion matrix was normalized which is illustrated in Fig. 8. It reflects the relative proportions of predicted labels compared to the true labels, providing insights into the overall effectiveness of the classification model. The confusion matrix demonstrates the system's high sensitivity in detecting mpox. While using HAM10000 weights led to increased accuracy for most architectures, it was not always advantageous for models. This may be due to the fact that the HAM10000 dataset solely consists of dermoscopic images, which is not the case for our dataset. For further qualitative analysis, we have generated heatmaps produced with Grad- CAM as depicted in Fig. 9. As shown in this figure, the best performing model can adequately

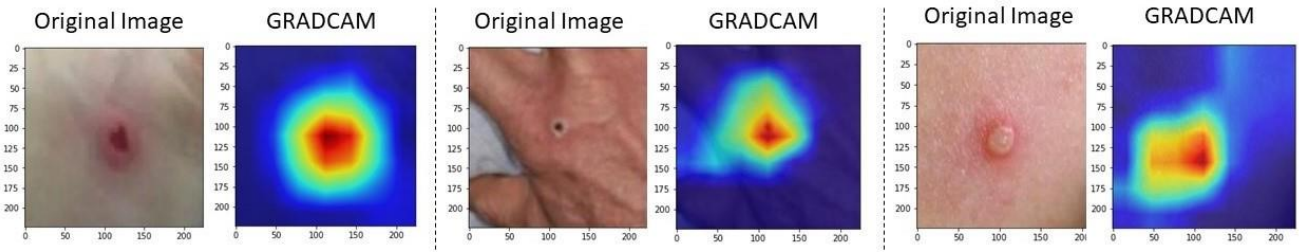

Fig. 9: Example skin lesion images and corresponding heatmaps produced via Grad-CAM using the best performing model.

localize and focus the skin lesions, demonstrating the interpretability of the model.

## VI. MPOX DETECTOR WEB APP

In this work, we have also developed an intuitive and user-friendly web application for online mpox skin disease detection to demonstrate our work. The web application is powered by the best-performing deep learning model presented in this work. The app's front end is built using HTML, CSS, and JavaScript. Upon clicking the upload button, users can easily upload a skin image using their phone's native camera app. With the user's appropriate consent, the uploaded data can be on our local server, which can be utilized for retraining the model for improved performance in the future. The prototype of the web application has been developed using the open-source Python streamlit framework with a flask core and has been hosted in the streamlit provided server for a better user experience. Fig. 10 shows the interface of the current web application. We plan to further improve the DL models and our web application in our future work. In the unfortunate event of another mpox outbreak, we believe such AI-assisted mpox screening tools can benefit disease surveillance and contact tracing.

## VII. DISCUSSION

In this work, we present our work on mpox skin lesion data collection and experimental evaluations of DL-based classification methods with an aim to overcome some of the shortcomings of the previous works in this area. First, we developed an updated version of our dataset, MSLD v2.0, including a few additional classes of infectious skin diseases and additional images for the existing classes of measles, chickenpox, and mpox. The dataset includes 755 images of skin lesions of 541 distinct patients. These additional images improve the models' capabilities to generalize new cases. Second, in the previous version of our dataset, there was an under-representation of dark skin tone images, possibly affecting the classification algorithm. We used color space augmentation, which reduces racial and regional biases in the dataset, to address this problem. Finally, in our previous works, the pre-trained weights from the ImageNet dataset, which essentially comprises all types of images, were utilized for transfer learning. In our current work, we used pre-trained weights in the HAM10000 dataset containing 10,015 dermoscopic pictures to do transfer learning. This strategy improved the model's accuracy, as discussed in the previous section. However, large-scale mpox data collection efforts are still required to develop more generalizable models for mpox disease screening. Our dataset was primarily compiled via web-scraping and thus lacks crucial meta-data, such as the patient's clinical history, the duration of the sickness, and the stage of the disease, which are essential for diagnosis. A more coordinated effort and international collaboration are required to build a larger dataset that can provide generalized results across different demographic regions.

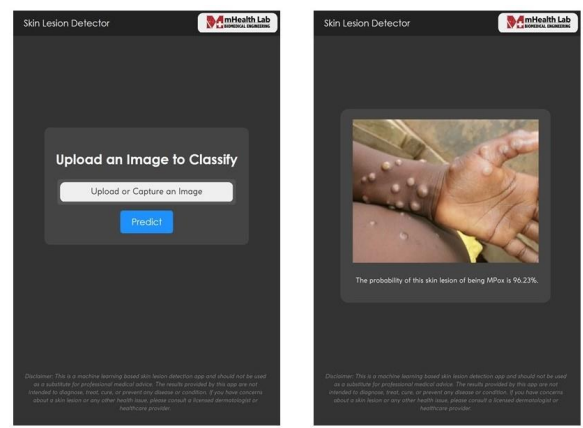

Fig. 10: The user interface of the online mpox screening tool.

## VIII. CONCLUSION

This study has presented an open-access dataset and a promising approach for automatically detecting mpox from skin lesions using state-of-the-art deep learning architectures. The "Mpox Skin Lesion Dataset (MSLD) v2.0" presented in this study can be beneficial to researchers to advance this field of research further. The experimental results demonstrate the potential and effectiveness of deep learning-based AI systems on this dataset for early diagnosis of mpox and other infectious diseases. In addition, we have also developed a web application that can play a significant role in public health by allowing people to conduct preliminary screening during the early phases of infection. Future works can focus on expanding the dataset by incorporating data from diverse geographical locations worldwide, reducing the under-representation of particular ethnic groups, and improving the generalizability of the models. In addition, further research on lightweight DL models will enhance the efficiency and ease of deployment in edge devices, improving public access to such AI-assisted screening tools. We hope our current efforts will contribute to developing effective AI-powered infectious disease detection and screening systems.